\documentclass{article}

\usepackage[utf8]{inputenc}

\usepackage[utf8]{inputenc} 
\usepackage[T1]{fontenc}    
\usepackage{hyperref}       
\usepackage{url}            
\usepackage{booktabs}       
\usepackage{amsfonts}       
\usepackage{nicefrac}       
\usepackage{microtype}      
\usepackage{lipsum}
\usepackage{amsmath,amssymb}
\usepackage{bbm}
\usepackage{graphicx}
\usepackage{url}
\usepackage{ dsfont }
\usepackage{mathrsfs}
\usepackage{soul}
\usepackage{graphicx}
\graphicspath{ {./images/} }
\usepackage{ mathrsfs }

\def\E{\mathbb E}
\def\H{\mathcal H}
\def\R{\mathbb R}
\def\N{\mathbb N}

\def\F{\mathscr F}

\title{Learnability Can Be Independent of ZFC Axioms: Explanations and Implications}

\author{
  (William) Justin Taylor\\
  Department of Computer Science\\
  University of Wisconsin - Madison\\
  Madison, WI 53715 \\
  \texttt{wtaylor8@wisc.edu} \\
}

\begin{document}
\maketitle

\section{Introduction}
Since Gödel proved his incompleteness results in the 1930s, the mathematical community has known that there are some problems, identifiable in mathematically precise language, that are unsolvable in that language. This runs the gamut between very abstract statements like the existence of large cardinals, to geometrically interesting statements, like the parallel postulate.\footnote{Which states that, in a plane, given a line and a point not on it, at most one line parallel to the given line can be drawn through the point.} One, in particular, is important here: The Continuum Hypothesis, CH, says that there are no infinite cardinals between the cardinality of the natural numbers and the cardinality of the real numbers.

In Ben-David et al.'s "Learnability Can Be Undecidable," they prove an independence\footnote{In their paper, Ben-David et al. use the word "undecidable" to mean "statements not provable from the axioms of ZFC". I will use "independent" for that purpose and reserve "undecidable" to mean "statements a Turning machine cannot provide an answer to in finite time."} result in theoretical machine learning. In particular, they define a new type of learnability, called Estimating The Maximum (EMX) learnability. They argue that this type of learnability fits in with other notions such as PAC learnability, Vapnik's statistical learning setting, and other general learning settings. However, using some set-theoretic techniques, they show that some learning problems in the EMX setting are independent of ZFC. Specifically they prove that ZFC\footnote{Zermelo –Fraenkel set theory with the Axiom of Choice} cannot prove or disprove EMX learnability of the finite subsets on the [0,1] interval. Moreover, the way they prove it shows that there can be no characteristic dimension, in the sense defined in 2.1, for EMX; and, hence, for general learning settings.

Here, I will explain their findings, discuss some limitations on those findings, and offer some suggestions about how to excise that undecidability. Parts 2-3 will explain the results of the paper, part 4-5 will discuss some limitations and next steps, and I will conclude in part 6.

\section{EMX Learning}
\subsection{Searching for a General Model}
EMX learning is a newly created form of learning, and it exists for two purposes. Before we go into the specifics, it pays to understand them both.

First, it shows that our intuitive concept of learning has some gaps. That is, there are some setups where we, provably, cannot know whether a certain learning function will successfully learn.

Second, it shows that a general characteristic dimension for learning is not possible. For the last few decades, we have known that VC dimension characterizes PAC learning. Something is (eventually) PAC learnable iff it has a finite VC dimension. Since then we have found various other dimensions that characterize other settings: the Natarajan dimension characterizes multi-class classification, the fat-shattering dimension characterizes learning for real valued functions, etc.

\begin{center}
 \begin{tabular}{||c c||} 
 \hline
 Setting & Dimension\\ [0.5ex] 
 \hline\hline
 PAC Learning & VC   \\ 
 \hline
 Multi-class & Natarajan  \\
 \hline
 Real Valued Functions & Fat Shattering \\
 \hline
 ... & ...  \\
 \hline
 EMX & ???  \\ [1ex] 
 \hline
\end{tabular}
\end{center}

Many people working in the field had high hopes that we could define a general learning setting that encompassed all these individual settings and that we could find a characteristic dimension that resolved to the individual dimensions in each case. This could, presumably, give us a mathematically precise definition of learning. Moreover, hopefully, we would be able to fit more or less any particular machine learning algorithm into the general setting and get helpful bounds for its success.

Our EMX setting, however, will provably not have a dimension. So, assuming we can prove it fits into any reasonable general learning setting, we have proven that there is no possible general characterization.

\subsection{Finite EMX Learning}
First, we need to define EMX learning. Ben-David et al. begin with a motivating example. Suppose we are trying to find the best possible ad to display on a website. We might model the situation as such: There exists a set of people who might come to the website, and every person has a set of ads they would like. For any given person, we know what ads they like, but we do not know who will visit the site. So, for instance, consider the table below:

\begin{center}
 \begin{tabular}{||c c c c c||} 
 \hline
 Visitors & Ad 1 & Ad 2 & Ad 3 & ... \\ [0.5ex] 
 \hline\hline
 1 & 1 & 0 & 0 & ... \\ 
 \hline
 2 & 0 & 1 & 1 & ... \\
 \hline
 3 & 0 & 1 & 1 & ... \\
 \hline
 4 & 1 & 1 & 0 & ... \\
 \hline
 ... & ... & ... & ... & ... \\ [1ex] 
 \hline
\end{tabular}
\end{center}

Each row represents a person, and the "1"s represent which ads they like. Each column represents an ad, and the "1"s show which people it appeals to. 

We get a finite training set of people that visited the site, and we choose an ad. Our loss function is 0 for any person that visited the site and liked the ad, and 1 for any person who visited the site, but did not like the ad. There is no penalty for other people, who did not visit the site, liking the ad. As such, the obvious training algorithm is to find the column containing the most people in the training set.

However, because we will be using set theoretic techniques later, we will change the notation slightly. Instead of working in matrices, we will say that each ad is a set of possible visitors to the site. For instance, $ad_1 = \{p_1, p_4\}$, $ad_2 = \{p_2, p_3, p_4\}$, $ad_3 = \{p_2, p_3\}$, etc. Now, our problem is to receive a set of users and find the ad (set) that covers the most visitors.

\subsection{Generalized EMX Learning}
After setting up the motivating example, we generalize it. We wish to replace "website visitors" with arbitrary sets, and "ads" with propositions that are true about members of those sets. So, formally:

Let $X$ be a set, let $P$ be a probability distribution on $X$, let $\F$ be a family of functions $X \mapsto \{0, 1\} $. A learning function $G: \cup_{k \in \N}X^{k} \mapsto \F $ is an $(\epsilon, \delta)$ learner for $\F$ if, for all $\epsilon$ and $\delta$,  there is some $d \in \N$ such that:

$$Pr_{S \sim P^d}[\E_P G(S)) \leq \sup_{h \in \F} \E_P(h) - \epsilon] \leq \delta$$

Although $\F$ is technically a set of functions, we will almost exclusively discuss $\F$ as though it is the set consisting of objects mapped to $1$. This simply works better in all the proofs that follow.

Now, this definition immediately runs into a problem: Given the Axiom of Choice,  the standard topology on $\R$, and a standard probability space with Lebesgue measure, there are, provably, some non-measurable subsets of $\R$. As such, we need a way to confirm every member of $\F$ and every possible $G(S)$ is measurable, or $\E_P$ will not be well-defined. Moreover, we could end up with non-measurable functions, which could jeopardize our proofs later. As such, the Ben-David et al. implicitly assume\footnote{These assumptions have been confirmed through private communication.} the following three stipulations:

1. Every set we will use always has the discrete topology, so every subset of every set is open.

2. Our Concept Class is the set of all finite subsets of $X$; i.e. only finitely many objects from $X$ will be labeled $1$. And, the unknown distribution, $D$, assigns its entire probability mass to members of the concept class. So, only points labeled $1$ will be sampled.

3. Our probability measure for any set is to simply add the probability mass of each point in the set. Because of the previous point, this will always be a valid measure.

So, clearly under these stipulations $\E_P$ is well-defined; and, given 3, every possible function will be measurable. Thus, we have addressed our first set of concerns; however, there are some reasonable quibbles with these stipulations. 

First,  it is odd to insist on a discrete topology. This one, however, will take some time to address, so we will come back to it in a later section.

Second, philosophically, there is not a consensus over whether "learning" is the same as having "no regret". Above, we allow high error if our error is sufficiently close to $ sup_{h \in \F} \E_p(h)$. If this result only applied to no regret learning, that might be a real demerit. Luckily, however, that is not the case. We could just as easily have defined EMX learning without that term. The set of learnable things would decrease. However, we would still be able to reach the independence result.

Third, once we have fixed a concept class, most definitions of learning allow all distributions over $X$. Here, we only allow distributions with probability mass concentrated on the target concept. We will never see any negative examples. Formally, we know that for $Z \subset X$ such that $P(Z)=1$, for all $z \in Z$, $P(1|z)=1$. But, while this may be odd, it is the conceit making this form of learning unique, and this form can be both challenging and useful.

To be challenging, we have to set things up properly. First, we must have a proper learner: one that has to output a hypothesis from $\F$. If not, we can simply output the  "all 1's hypothesis," which outputs one for every element in the domain. Then, that solution is trivially optimal because our loss function does not penalize us for guessing $1$ when the correct output is $0$. Similarly, even with a proper learner, things could be trivial. If there is an $f \in \F$ that contains all the other hypotheses, that is also trivially optimal. Contrarily, suppose we are only allowed to label a finite amount of things $1$. Then it may still be quite challenging to find the best hypothesis.

There are also real life situations where this set up makes sense. We have already mentioned the website setup Ben David et al. describe. In addition, this set up might be useful in a cryptology set up. Suppose you have intercepted a small amount of a noisy message. You might use this method to predict future noise so that you can decode the rest of the message.\footnote{Throughout the paper, we view our task as discrimination. We are predicting something outside our control. However, for those who still feel uneasy about calling this set up learning, we could view things through a generative lens. We are, from this point of view, generating new, successful instances before seeing them ourselves. Generative modes are sometimes viewed as a model of the conditional probability of the observable X, given a target y, symbolically, $P(X|Y =y)$ which matches what we are doing more closely.}

\subsection{EMX as an Instance of a Generalized Learning}
Lastly, we want to ensure our definition lands inside reasonable definitions of a general learning settings. And, fortunately, it seems that it does. Like PAC learning, or regression estimation, we have a hypothesis space, an unknown distribution $P_{x,y}$, and our goal is to get sufficiently close to the best hypothesis in the hypothesis space. In fact, at first glance, it appears to meet the definition of learning in Vapnik's general learning setting:
\begin{quote}
The model of learning from examples can be described using three components:
1) a generator of random vectors $x$, drawn independently
from a fixed but unknown distribution $P_x$;
2) a supervisor that returns an output vector $y$ for every
input vector $x$, according to a conditional distribution
function $P(y|x)$, also fixed but unknown;
3) a learning machine capable of implementing a set of
functions $f(x,\alpha), \alpha \in \Lambda$. The problem of learning is that of choosing from the given set of functions $f(x,\alpha), \alpha \in \Lambda$, the one which predicts the supervisor’s response in the best possible way \cite{Vapnik 1999}.  
\end{quote}

The first two conditions are immediately met, and the third is questionably met. Obviously, "best possible way" is rather vague, but we will take it to mean minimizing the expected loss. In that case, if our learning function, $G$, counts as a learning machine, then we clearly do meet it. The function, $G$, we develop in the next section will repeatedly apply the Axiom of Choice, so it is not an algorithm. But, recent literature\footnote{See for example \cite{Simon1997}} has moved to simply discussing learning functions, instead of learning algorithms. If that move is legitimate, then we have succeeded.

\section{Tying Learning to Compression}
\begin{figure}
    \centering
    \includegraphics[scale=.5]{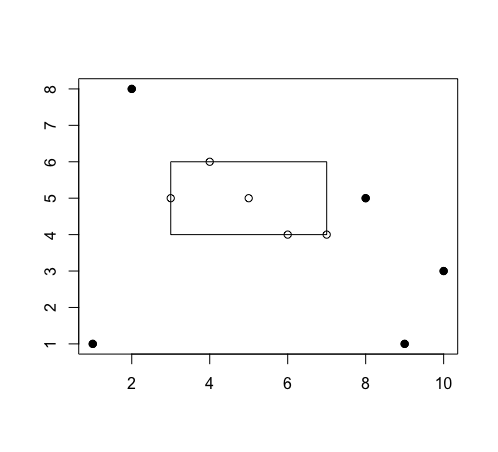}
    \caption{We see that we have compressed our ten data points down to four. Moreover, any future data points would be similarly compressed.}
    \label{Pictoral Representation of Recangle Compression}
\end{figure}
In this section, we will see how to use set theoretic compression techniques to achieve the our desired learnability results. Littlestone and Warmuth's classic paper, "Relating Data Compression and Learnability"\cite{Littlestone1986}, did just what its title suggested. Suppose there exists $d \in \N$, and functions $\sigma, \eta$, such that for all $M \geq d$, $\sigma$ can compress $M$ labeled, test points down to $d$ ones,
then use $\eta$, necessarily measurable, to classify all points in the domain, using only information in the $d$ points, that assigns all the same values to the points which were compressed out. In that case, the underlying concept is learnable for $(\epsilon$ and $\delta)$, as a function of $d$. Moreover, later papers proved the reverse, if there is no possible compression, then the concept is not learnable.\cite{David2016}\cite{Moran2016}

\subsection{Defining Our Compression}
\subsubsection{Intuitions}
Before we get into the specifics of our compression and decompression functions, we should discuss some motivating examples. Suppose our hypothesis class, $H$ is all axis aligned rectangles in $\R^2$. True examples are inside some rectangle, and false examples are outside that rectangle. Suppose we get test sample $S$. Then we could define the following compression and decompression functions:

$\sigma$: Find $s_1, ..., s_4$ such that $s_1$ and $s_2$ had the least and greatest horizontal values, and $s_3$ and $s_4$ have the least and greatest vertical values.

$\eta$: Draw a rectangle whose horizontal boundaries go from that of $s_1$ to $s_2$ and whose vertical goes from $s_3$ to $s_4$. Everything inside is true, and everything outside is false.

Clearly, any point in $S$ will still be labeled the same, and we now have assignments for all values in $\R^2$. So, we have a valid compression learner. See Figure 1 for a pictorial representation.

Although a bit too complicated to go through here, this is the same idea behind (hard margin) support vector machines. The algorithm first finds the unique data points closet to those with an opposite classification, ignores all the other data points, then creates a margin that classifies the whole space in such a way that all previous test points will have the same value.

Finally, here is a third example that very closely parallels the compression we will define in 3.1.2. There is a famous problem from World War II called the German Tank Problem. The Allies want to know how many tanks the Germans have produced. They have captured a few German tanks, and they know that Germans have issued sequential serial numbers for their tanks.

One simplistic strategy would be to assume that they have produced exactly the number of tanks listed on the highest serial number you have collected so far. Clearly, this is asymptotically correct. And, it is easy to see what your expected error should be. If you implemented this strategy, then you have effectively compressed all your data into one data point.

So, we see that compression based learning is often successful. And, because of Littlestone and Warmuth, we know that if there is any learner, then there is also a compression one. We have now arrived at Ben David et al.'s original results, which will be explained throughout the rest of section 3.

\subsubsection{Definitions}
We need to define a $d, \sigma$, and $\eta$ for our EMX learning problem. However, it turns out that it is easier to define $\sigma$, and $\eta$ as $m+1 \rightarrow m$ and $m \rightarrow m + n$ functions. As such, we will first define them that way, and then extend them to an $M \rightarrow d$ compression.

A monotone compression scheme is a pair of functions, $\sigma$ and $\eta$, such that $\sigma:$ $m+1$ subsets of $X$ $\rightarrow $ $m$ subsets of $X$ and $\eta:$ $m$ subsets of $X \rightarrow P(X)$ such that for every $m+1$ subset in $P(X)$, $\beta$, that subset is contained in $\eta(\sigma(\beta))$.

In our case, we define a compression scheme from $k+2$ to $k+1$ for any $X$ such that $|X|= \aleph_k$ through the following procedure. Let $|X|=\aleph_k$, and let $S \subseteq X$ be of size k+2. Let $\prec_k$ be a well-ordering of X of order type $\omega_k$. Note that, because the well-ordering theorem is equivalent to the axiom of choice, $\prec_k$ is guaranteed to exist. Take the maximal element of S in this ordering. The initial segment $X$, under $\prec_k$, has a cardinality of, at most, $\aleph_{k-1}$. Get a new well-ordering $\prec_{k-1}$ of that initial segment and repeat with the maximal element on that new ordering of the remaining elements of $X$. Repeat until you get to $\aleph_0$. Take the greater element, $s'$, of the remaining 2. The last element $s^*$ will be compressed out. Completing our definition of $\sigma$. See Figure 2 for a pictorial representation.

From there, our definition of $\eta$ is obvious. $\eta(S/s^*) = S/s^* \cup \{$The finite set of points less than $s'\}$. The latter will, necessarily, include $s^*$, so the compression is sound. Clearly, this also gives us our $d$. $d$ is simply $1$ greater than the index of the cardinality of the domain.

\begin{figure}
    \centering
    \includegraphics[scale=.5]{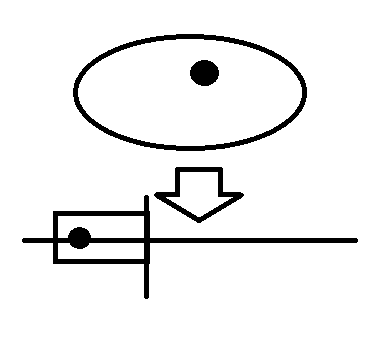}
    \caption{Suppose $|[0,1]|=\aleph_1$. Then we take a point in our $\omega_1$ order, get a set of $\aleph_0$, choose some maximum, and compress out the remaining point. Our reconstruction is to choose all points less than that second maximum point.}
    \label{Pictoral Representation of our Compression}
\end{figure}

\subsubsection{Generalizing to any Amount of Training Data}
Finally, now we can return to getting an $M \rightarrow d$ compression. It turns out that we simply do the intuitive thing. Given $M$ points, take some $d+1$ of them and compress them to $d$ points. From those $M-1$ points, take $d+1$ points and compress that to $d$ points. Repeat this procedure a total of $M-d$ times until you have $d$ points. To decompress, simply apply $\eta$ on the $d$ points. Then apply $\eta$ on every $d$ subset of the result. Then repeat until you have completed $M-d$ iterations. The result is guaranteed to be finite, since every step added finitely many elements, and it was done a finite number of times. It is also guaranteed to contain every original element. Since, at some point, we applied $\eta$ on exactly the $d$ elements that compressed that element out.

\subsection{Proving No Other Form Of Data Compression Exists}
Of course, while we have just shown that there is a compression for $|X| < \aleph^{\omega}$, we have not yet shown there there may yet be some other way to compress data for sets of higher cardinalities.

First, clearly, it suffices to show that higher cardinalities have no monotone compression schemes. For, if there is some other compression scheme that compressed points from $a \rightarrow b$, for, $a >> b$, then we can turn it into an $a \rightarrow a-1 $ monotone compression scheme by simply compressing down to $b$ points, then adding all but one of the points back in.

So, the Ben-David et al. focus on monotone compression schemes and use a proof by infinite descent to prove the following result:

{\em Lemma} Let $k$ be a positive integer and let $Y \subset X$ be infinite sets with cardinalities |Y| < |X|. If $\F^X_{fin}$ has a (k+1) $\rightarrow$ k monotone compression scheme then $\F^Y_{fin}$ has a $k \rightarrow (k-1)$ monotone compression scheme.

This lemma completes our proof because it implies no set bigger than $\aleph_k$ can have a $(k+2) \rightarrow (k+1)$ compression scheme. After all, there is obviously no $1 \rightarrow 0$ compression scheme for $\aleph_0$. The rest follows by induction.

{\em Proof of Lemma}
Let $X$ and $Y$ be as above. Let $\sigma$ and $\eta$ be the compression and decompression functions guaranteed by $X$'s compression scheme. Consider the union of all elements in $\eta(Y)$, $Z$. Since $\eta(y)$ is finite for any particular element in $X$; and, therefore, in $Y$, $|Z| = |Y|$. So, there is some $x \in X / Z$. Then we can build a $k \rightarrow (k-1)$ compression scheme as follows: Let $s \in Y^k$. Take $\sigma(s \cup {x})$. This is guaranteed to still contain $x$. If it didn't, $x$ would be in $Z$. So, $\sigma(S \cup {x}) / {x}$ is our desired $k -1$ set. Since this works for an arbitrary subset this provides our $\sigma_{Y}$, and $\eta_{Y}$ immediately.

\subsection{Independence from ZFC and Failure of Characteristic Dimension}
With all this work in place, it remains to conclude that learnability is independent of ZFC. We have learnability iff we have compression, and we have compression iff our domain is $\aleph_k$ for some $k \in \N$. If CH is true, then $|[0,1]| = \aleph_1$. If it is false, then it might be that $|[0,1]| > \aleph_k$, for all $k$. Since the latter is independent of ZFC, so is the former.

Finally, Ben David et al. suggest that dimensions that characterize different learning settings must be of finite character. To quote them: "A property $A (\mathscr{X},\mathscr{Y})$ is a finite character property if there exists a bounded formula $\phi (\mathscr{X},\mathscr{Y})$ so that ZFC proves that A and $\phi$ are equivalent." Intuitively, these are properties that can be checked by probing finitely many elements of $\mathscr{X}$ and $\mathscr{Y}$. They note that every working dimension of learning is of finite character, so this seems like a fair constraint.

However, any bounded formula has the same truth value in any model of ZFC. Hence, every finite character property must have the same truth value in all models of ZFC, as well. Since any dimension that accurately characterized EMX learning could not have the same truth value in all models of ZFC, Ben David et al. conclude that there must be no finite character property for EMX.


\section{Trough of Disillusionment}
When discussing new technology, there is a graph called the Hype Cycle. It begins with the Peak of Inflated Expectations, where we have just learned of the technology, and we think it is revolutionary. That is analogous to where we are now. Ben-David et al.'s results seem fantastic. It seems that, simply through well-ordering the reals, they have found an interesting learning problem that cannot be solved. 

This seems rather worrying. However, there are a few reasons this result is not as powerful as it first appears. Hence, the name of this section. Following the Peak of Inflated Expectations is the Trough of Disillusionment where I will explain why these results are not quite as powerful as they appear.
\begin{figure}
    \centering
    \includegraphics[scale=.5]{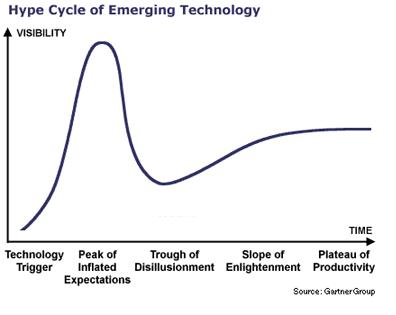}
\end{figure}

\subsection{Topological issues}
In section 2.2, I mentioned that all sets have the discrete topology. There, I noted that this ensured our expectation function was well-defined, and it ensured that our functions were well-defined. Here it is time to expound on the latter claim.

In the specific set up used to show undecidability, there are much more natural topologies. Consider that all three sets are composed of elements of $[0,1]$ of various sizes. First, $d+1$, then $d$, then allowed to vary. These can easily be represented as $[0,1]^{d+1}$, $[0,1]^{d}$, and $[0,1]^{\omega}$. 

These representations are trivial in the first two cases. Each set containing $\{x_1,...,x_d\}$ is associated with $(x_1,...,x_d)$, and we ensure rearrangements of the elements all have the same image under $\sigma$ and $\eta$. 

For, $[0,1]^{\omega}$, we associate $\{x_1,...,x_m\}$ with $(x_1,...,x_m, 0,..)$. So, the first $m$ elements are the members of the set, and there are countably many zeros after that. We will insist that the first $m$ elements are ordered under the natural ordering, to distinguish  $\{x_1,...,x_m\}$ from  $\{x_1,...,x_m, 0\}$.

With this in hand, we can assign natural topologies to these three sets. Namely, balls of the appropriate dimension. If we do so, things immediately break down. Here, I am going to simply list several theorem statements; the proofs are all in the appendix.

\textbf{Theorem 1}: $\sigma:[0,1]^{d+1} \mapsto [0,1]^d$ with the natural topology on both sets is not Borel measurable.

\textbf{Theorem 2}: $\eta:[0,1]^{d} \mapsto [0,1]^{\omega}$ with the natural topology on both sets is not Borel measurable.

Moreover, I recognize that $[0,1]^{\omega}$ is potentially a less natural fit then the first two, hence I think it's also important to explore what happens when we leave it, as is, in the discrete topology. However, this doesn't change things.

\textbf{Theorem 3}: $ \eta:[0,1]^{d} \mapsto \F^{[0,1]}_{fin}$, where $[0,1]^{d}$ has the natural topology and $\F^{[0,1]}_{fin}$ has the discrete topology is not measurable.

Collectively, these three theorems show two fairly significant things: 

First, as mentioned in \cite{Muller 2007}, since our learning function is $G(x) = \eta(\sigma(x))$, and since compositions of finite to one functions that are not Borel measurable are still not Borel measurable, under reasonable topologies, $G$ is not measurable. This immediately disqualifies $G$ from ever being implemented in the real world. No algorithm could ever implement such a function.

Second, as I mentioned in the beginning of part 3, Littlestone and Warmuth’s theorem only guarantees that a compression scheme implies learnability when the reconstruction is measurable. Hence, if we were to insist that $[0,1]$ were to have all of its normal structure, $\eta$ would not be measurable, and the entire proof would fall apart.

\subsection{Entanglement with the Axiom of Choice}
Setting the measurability issues aside, the independence result also has an interesting dependence on the Axiom of Choice. There's nothing inherently wrong about that. The authors very explicitly state they were using ZFC set theory in this paper; and, of course, if learnability is independent of ZFC set theory, it is certainly independent of ZF set theory. However, the mathematical community has skepticism for the axiom of choice in a way it does not for other ZF axioms, making this dependence interesting.

A standard definition for this axiom is: for any set $X$ of nonempty sets, there exists a choice function, $f$ defined on $X$, where a choice function, $f$, with domain $X$ is such that $f(A) \in A$, $\forall A \in X$. This sounds relatively harmless; and, indeed, true. Certainly if we have a collection of sets, we can "pick" a member from each of them, for a new set. There are other equivalent statements that also seem "obvious". See, for instance, Zorn's lemma or Trichotomy for cardinality. It also has results that are less obvious, but very exciting. For instance, it is equivalent to statements such as the well-ordering theorem and that every vector space has a basis.

However, the axiom of choice also proves some particularly shocking results. The most famous is the existence of non-measurable sets, and the Banach-Tarski Paradox\cite{Tao2011}. Another appears in a generalization of white hat/black hat riddles\cite{Muller 2007}. Moreover, one might think that ML theorists, in particular, might want to avoid appeals to the Axiom of Choice. If it is impossible to create a choice function without an appeal to this axiom, then no algorithm can take advantage of such a choice function. So, it certainly seems questionable to say something is learnable on the basis of that choice function. So, if the Axiom of Choice is not required, that would put this result on firmer footing.

Before proceeding, I think it pays to go through the logical implications slowly, since there is some ambiguity. Let ZF\footnote{The symbolic notation in this paragraph follows Kunen\cite{Kunen 1980}}, ZFC, and ZF$\neg$C, be the the rest of the Zermelo-Frankel axioms omitting the Axiom of Choice, with the Axoim of Choice and with the negation of the Axiom of Choice. Let U stand for the statement that some EMX learning problem is undecidable, let I stand for the Interval being EMX learnable, and I* be the statement that the interval is undecidable in EMX. I've already mentioned that if I is independent of ZFC, it's independent of ZF. That means, if ZFC + I and ZFC + ($\neg$I) are consistent, ZF + I and ZF + ($\neg$I) are too. However, that doesn't preclude that if the interval is (provably) independent, then the axiom of choice it true. I.e. ZF + I* $\rightarrow$ C. Moreover, even if the interval's independence does depend on the axiom of choice, there may be some other set which doesn't. I.e. even if ZF + I* $\rightarrow$ C, ZF$\neg$ C + U might be consistent. Finally, while any of these things might imply a a choice function on some particular sets, they might not imply that there is a choice function for every possible set.

So, with those subtleties in mind, we proceed. Our goal for the rest of the section will be to (I) recount the independence result's dependence on the axiom of choice and (II) show that the independence result, in turn, implies a weaker form of the Axiom of Choice.

(I) is actually extremely easy. One might say $G$ is doing nothing but repeated applications of the Axiom of Choice. At each stage is gets a well-ordering of a subset of $[0,1]$, exploits it to find the greatest value, creates a smaller subset and repeats. We only know the subset's well-ordering exists because of the well-ordering theorem, and the well-ordering theorem is equivalent to the Axiom of Choice.

(II) is a bit more difficult. Since all the proof does is iteratively assert the existence of a well-order and then exploit it, it feels like there should be a simple way to reverse the proof and use the existence of a compression scheme to prove the existence of a well-order. This would then show that every set, at least every set that is $\aleph_k$ for some k, is well-ordered. Unfortunately, this doesn't quite work. There's no reason that $\eta$ can't create loops. I.e., we compress $\{$.1, .2, .3$\}$ to $\{$.2 .3 $\}$ but $\{$.1, .2, .4$\}$ to $\{$.1 .4 $\}$, which means that they don't directly induce a well-ordering.

Fortunately, there is a way around this. We have already shown that, if we substitute in normal topologies, $G$ is not a Borel measurable function. However, it is consistent with ZF that $\R$ is the countable union of countable sets, which would imply that all reals were Borel. If we have a function that is not Borel measurable, that is no longer the case. So, there must be a choice function on [0,1].\footnote{To be precise, there need only be choice functions on subsets of [0,1] with cardinality $\aleph_k$ for some k. In the case where the cardinality of [0,1] is bigger than that, the larger sets would not provably have such a choice function.} Note that we haven't quite proved the full axiom of choice. There might be some other set in the universe of sets, bigger than [0,1] that has no choice function. But, we did prove one for the interval.

\subsection{A Clearer Representation of The Problem We Solved}
So, a lack of some of $[0,1]$'s structure is necessary for this theorem to hold. The only structure actually used in the proof, which is also provably necessary for the theorem's truth, is a well-order. This shows that we aren't \emph{really} talking about $[0,1]$, at all. We are actually dealing with ordinals.

Consider our German tank problem from before. There we compressed all of our data into only the greatest serial number. Here we are doing almost the same thing. We take advantage of the theorem that any initial segment of a limit ordinal is of a smaller cardinality. Then, instead of compressing once into the greatest element, we compress into the $d$ greatest elements to traverse the distance from $\aleph_{d}$ to $\aleph_{0}$ where we can get the greatest of the finite set remaining. 

Other than the multiple steps, it is identical; and the ordinals are even provably well-ordered without any appeal to the Axiom of Choice. 

In fact, if we had a magic machine that could take any two ordinals less than $|2^{\aleph_0}|$ as input and return the greater, then the rest of this learning setup could be implemented quite trivially. Ironically, it would provide is with an empirical experiment on the truth of the Continuum Hypothesis in our universe.

But, put this way, the result is a bit less exciting. It seems less like the authors found a gap in the concept of learnability that needs to be closed and more like the authors found that our definition technically applies in a case we do not normally care about. The first seems like a case where we need to tighten our definition. The second seems more like a case where we shrug our shoulders and conclude that, \emph{technically,} there are learning problems where learning depends on axioms outside of ZFC.

Of course, they have still shown that not every learning setting can have a characteristic dimension. So, even if we ignore the philosophical concerns we will want to hunt for nonrestrictive properties we can use that would always have a characteristic dimension.

\section{Restricting What Can Be Learnable}
In this section I offer two ways we could go about restricting learnability. Neither will provably necessitate a characteristic dimension, but both do a good job of solving EMX independence without creating undue burden elsewhere. They are: (1) requiring that our learning function be implemented as an algorithm and (2) requiring our domain be countable. However, in a third section, I will note that while either neatly resolves the independence from Ben David et al.'s paper, it cannot solve either of our problems in full generality. We have no proof that (1) and (2) ensure a finite characteristic dimension; and, even with (1) and (2), there are still some other problems that would be independent of ZFC.

\subsection{Restricting to an Algorithm}
In the paper, the authors argue that our difficulties come from "the existence of a learning function rather than the existence of a learning algorithm."\cite{David2019}. That is, we need to explicitly require that an algorithm can take the given examples for $P_{x,y}$ and return a particular hypothesis from the hypothesis class.

This does seem like it would neatly solve our problem. No algorithm could take advantage of the repeated well-orderings on subsets of [0,1], so no algorithm could EMX learn the distribution using that method. In fact, as Hart has shown any appropriate learning function would not be Borel measurable, it seems we have a proof that no algorithm would suffice. If we have a function that is not Borel measurable, then we needed to invoke the Axiom of Choice, and if we need the Axiom of Choice for a choice function, then an algorithm cannot implement that function. Hence, [0,1] is simply not learnable under this new definition.

So, how does this square with other types of learning? In Shavlev-Shwartz and Ben-David's {\em Understanding Machine Learning From Theory To Algorithms}, Agnostic PAC Learning is defined in terms of an algorithm already. So, there won't be any problems there. Moreover, at first glance, it seems this generalizes quite well. Here, we will recreate the $\Psi$ learning context since it's fairly general, but, Natarajan and Fat-shattering dimensions should be analogous:

\textbf{Learnability adopted from David 1995 with new text in bold}\cite{David1995}:
        [W]e say that $\F$ is \emph{learnable} if there exists {\em \textbf{an algorithm A, computable,}} and an integer-valued function $m=m_{\epsilon,\delta}$ such that for any $\epsilon,\delta > 0$, for any probability measure $P_{x,y}$ over $X$ and for any $f \in \F$, the event $error_{P_{x,y},f}(A(\bar v)) > \epsilon$ occurs with probability at most $\delta$ for random sequences $\bar v = ((x_1, f(x_1)), ... , (x_m,f(x_m)))$, where $(x_1,...,x_m) \in X^m$ is drawn according to $D^m$.

It seems like this is a viable path forward for the field. However, it comes with two downsides. The first is mentioned in the paper: it is simply easier to work with functions than to work with algorithms. Proofs about existence or non-existence of functions are simpler, and they separate information theory from algorithmic complexity theory. The second is that we may just be moving our vagueness problem from "learnability" to "algorithm". For instance, in Vapnik's definition of learning the "learning machine," our algorithm, should be capable of implementing any function in the function class $\F$. However, our algorithms often do not meet this requirement. To take a toy example, consider the class of threshold classifiers on [0,1]. No algorithm could implement a threshold at any non-computable number. Since the computable numbers are only countable, this means that the ones we could implement are of measure zero. Surely, there is a way to correct this problem. For instance, perhaps we just insist the algorithm implement something arbitrarily close to any member of the hypothesis class. But, it is non-trivial, and it leads neatly into another way to solve this problem.

\subsection{Restriction to a Countable Domain}
When we, the ML community, design algorithms, we commonly talk about our feature space as sets like $\R^{d}$. However, omitting a few necessary exceptions, such as $\pi$ and $e$, we almost never actually enter any transcendental numbers into training or learning sets. And, of course, the same is true of our algorithm's output.

We would never want to limit ourselves to only the binary strings of numbers representable on this or that computer, since that would make our results relative in a way we do not want. I do, however, think we can cut out the numbers that aren't even representable, in theory. As such, I propose we instead define our hypothesis classes to have domains and ranges over subsets of the computable reals.

Informally, "a computable number [is] one for which there is a Turing machine which, given $n$ on its initial tape, terminates with the $n$th digit of that number [encoded on its tape]."\cite{Minsky1967}\footnote{This definition is subject to a problem called the table-makers dilemma. So, there is a modern, more technical definition. But, the distinction is not important here.}. Clearly, this is the set we are looking for. If there is no Turing machine that could be counted on to give some number to our algorithm, then there is no way for our algorithm to meaningfully use it, nor to output it to us. So, we are not artificially constraining ourselves in any meaningful way.

Also, notice that we immediately get an obvious solution to our undecidability problem. There are now only countably many numbers in $[0,1]^{computable}$. There clearly is a 2-1 compression scheme here. You just create the list of those numbers. Get the greater number on that list, and output the finite subset of numbers less than or equal to the greater number.\footnote{Notice, that it's still not the standard ordering.} So, on this disambiguation of the problem, it is learnable.

Note, that this definition is meaningfully distinct from requiring an algorithm. For instance, in the case above, there is still no (computable) algorithm that could learn this. The computable reals are not computably enumerable. So, there is no computable function that can make the list above. It's unclear to me whether this is desirable. Though, if it isn't, we could simply insist that we both require an algorithm and for our domain and range to be computable. The solutions are in no way in conflict with each other.

The most interesting thing about this restriction is how it would affect proofs. The computable numbers are closed under most normal operations, so most of our basic results will still stand. However, they are not closed under suprema, so we will need to make some adjustments. For instance, disagreement coefficients cannot be assumed to be computable. As such, the CAL algorithm will have to be adjusted. This will not be a major problem because the ultimate goal is to choose an integer number of examples to check at each epoch. So, some finite level of closeness to the true $\theta \in \R$ should be fine. And, generally, this method should work for most other proofs.

I wholeheartedly do consider this a desirable feature of this method. If our proofs did truly depend on using an uncomputable number, exactly, and not a sufficiently close computable one, then this is a major problem. We could not actually program a computer with such a number.

\subsection{Continued Failure In Pathological Examples}
Now, obviously, we do not yet know whether either of these restrictions will yield a general setting that always has a characteristic dimension. That would take significant further research. However, we might hope that these restrictions, and disbarring using the Axiom of Choice, would collectively leave us unable to think up a learning problem that is still independent of ZFC. Unfortunately, even that is not true. In Yedidia and Aaronson's 2016 paper\cite{Yedidia2006}, they exhibit a Turing Machine, $BB(n)$ based off Rad\'o's Busy Beaver function. ZFC cannot prove whether or not this Turing Machine halts.

Take the most garden variety learning problem with hypotheses $\H$. Pick some specific model of ZFC where $BB(n)$ halts, and let $c$ be one greater than the point it halts in that model. Let $\H^*$ be those same hypotheses except that if $BB(n)$ has not halted in $c$ steps, they all output the all $0's$ function. This set up does not depend on the Axiom of Choice, since $BB(n)$ does not. It only deals with computable numbers, and it has a learning algorithm that can implement every member of $\H^*$, which are all algorithms. But, whether it learns is still independent of ZFC.\footnote{Note that none of this invalidates the VC dimension as a characteristic dimension of finite character for PAC learning. Our different models of ZFC vary the VC dimension. They do not vary the relationship between having a finite VC dimension and being learnable.}

Obviously, this set up feels illegitimate. Adding in $BB(n)$ suggests we are not really \emph{trying} to learn. But, the point is that while we should try to excise gaps in our definition that are in the way of advancing our theory, we can probably never excise all gaps without running the risk of making our new definition trivial.

\section{Conclusion}
Ben David et al. really have found a set up where learnability is independent of ZFC; but, as I mentioned in 4.3, this may be less disheartening then it first appears. While a result about the unit interval with all its normal structure might be a serious wake up call, we just might not care that compressing ordinals happens to meet our normal definition of learning and that ordinals have odd cardinal properties. They do clearly show that our current definition is too broad to ensure a general setting that always has a characteristic dimension. However, as 5.3 shows, it is very hard to excise all independence so we should be skeptical that any of our solutions will ensure such a dimension until the subsequent work is done.

\bibliographystyle{unsrt}

\appendix
\section{Proofs}
Lemma 1 and Theorem 1 both come directly from Hart. We will explicitly use the Lebesgue measure\cite{Hart 2019}.

\textbf{Lemma 1}: $\sigma:[0,1]^{d+1} \mapsto [0,1]^d$ with the natural topology on both sets is not continuous.

\emph{Proof}:
Suppose $\sigma$ is continuous. Let $x \in [0,1]^{m+1}$. Without loss of generality, $x_m$ is always compressed out; i.e. $\sigma(x) = \{x_0, ..., x_{m-1}\}$. Let $\epsilon = \frac{1}{3}min\{|x_i -x_j|: 0 \leq i,j \leq m-1\}$. Let $\delta \geq 0$ be such that $\delta \leq \epsilon$ and $\forall y \in  [0,1]^{m+1}$, if $||y-x|| \leq \delta$, then $||\sigma(y)-\sigma(x)|| < \epsilon$.

So, let $y \in [0,1]^{m+1}$ with $||y-x|| \leq \delta$. Clearly, $|y_i-x_i|| < \epsilon, \forall i \leq m$. Because of the triangle inequality, $y_m - x_i > \epsilon, \forall i < m$. So, $\sigma(y) = \{y_0, ..., y_{m-1}\}$. But, this is true for any $y$ of this character. Hence, it is true for infinitely many $y$, so $\sigma$ is not finite to one.

\textbf{Theorem 1}: $\sigma:[0,1]^{d+1} \mapsto [0,1]^d$ with the natural topology on both sets is not Borel measurable.

\emph{Proof}:
We assume  $[0,1]^{m+1}$ with the Lebesgue measure is a Radon measure space, and $[0,1]^{m}$ with the Lebesgue measure is a second countable topological space. So, by the general form of Lusin's theorem, there is a closed set $C$ with measure greater than $0$ such that $\sigma$ restricted to $C$ is continuous. But, from there, we are basically done. On this restricted, continuous function, the previous proof applies; so, again, $\sigma$ is not finite to one.

\textbf{Theorem 2}: $\eta:[0,1]^{d} \mapsto [0,1]^{\omega}$ with the natural topology on both sets is not Borel measurable.

\emph{Proof}
Let $\eta: [0,1]^d \rightarrow [0,1]^{\omega}$ where both have the natural topology. That is, we order the first $d$ points and then add  finitely many points to the end based on that well-ordering. This cannot be guaranteed to be measurable either.

There are $|P([0,1])|$ well-orderings on $[0,1]$. Each creates a different $\eta$. However, there are only $|[0,1]|$ Borel measurable functions.\cite{stack3} Since we choose our well-ordering arbitrarily, through the Axiom of Choice, we cannot ensure that ours is measurable.

\textbf{Theorem 3}: $ \eta:[0,1]^{d} \mapsto \F^{[0,1]}_{fin}$, where $[0,1]^{d}$ has the natural topology and $\F^{[0,1]}_{fin}$ has the discrete topology is not measurable.

This proof was adapted from a proof on math stack exchange. That was, in turn, adapted from a classic reconstruction of a nonmeasurable set.\cite{stack1}

{\em Proof}
Let the ordinal $c$ be the cardinal of $[0,1]$, which is also the cardinal of $[0,1]^m$ and the cardinal of the set of all closed uncountable subsets of $[0,1]^m$ and the cardinal of each uncountable closed subset of $[0,1]^m$.

Let $S=\{f^{-1}(s):\emptyset \neq s \in \F^{[0,1]}_{fin}\}$. Then, $S$ is a pair-wise disjoint family of finite sets and $\cup S = [0,1]^m$.

Let $\{C_a: a < c \}$ be the set of all uncountable closed subsets of $\R^m$.

Define $\{A(d): d < c \}$ and $\{B(d): d < c \}$ as follows, recursively:

For $a < c$ suppose that $\{A(d): d < c \}$ and $\{B(d): d < c \}$ are subsets of $S$. Then the cardinals of $\cup\{A(d): d < c \}$ and $\cup\{B(d): d < c \}$ are each at most $|a \times \omega |$which is <$c$. So, we may take distinct

$$x_a, y_a \in C_a / (\cup\{A(d): d < c \}) \cup (\cup\{B(d): d < c \})$$
such that $f(x_a) \neq f(y_a)$.

Then define $A(a) = \{f^{-1}(f(x_a)\} \cup \{A(d): d < a \}$ and $B(a) = \{f^{-1}(f(y_a)\} \cup \{B(d): d < a \}$

Now, let $A = \cup\{A(a): a < c \}$ and $B = \cup\{B(a): a < c \}$  Then, $A$ and $B$ are disjoint.

We have $A = f^{=1}\{f(y): y \in A\}$.

But $A$ is not Lebesgue-measurable because for every $C_a$ we have $x_a \in A \cap C_a$ but $y_a \in B \cap C_a \subset ([0,1]^m \ A) \cap C_a$.

Lebesgue measure, $m$, is inner-regular. If $A$ were measurable, then $m(A) = 0$ because the closed subsets of $A$ are countable. However, the only closed sets that are disjoint from $A$ are countable as well, so $m([0,1]^{m} / A) = 0$. This contradiction concludes are proof.

\end{document}